\documentclass[sigconf]{acmart}

\usepackage{colortbl}
\AtBeginDocument{%
  \providecommand\BibTeX{{%
    \normalfont B\kern-0.5em{\scshape i\kern-0.25em b}\kern-0.8em\TeX}}}

\copyrightyear{2022}
\acmYear{2022}
\setcopyright{rightsretained}
\acmConference[FDG '22]{FDG '22: Proceedings of the 17th International
Conference on the Foundations of Digital Games}{September 5--8,
2022}{Athens, Greece}
\acmBooktitle{FDG '22: Proceedings of the 17th International Conference
on the Foundations of Digital Games (FDG '22), September 5--8, 2022,
Athens, Greece}\acmDOI{10.1145/3555858.3555930}
\acmISBN{978-1-4503-9795-7/22/09}

\begin{document}

\title{Video Games as a Corpus: Sentiment Analysis using Fallout New Vegas Dialog}

\author{Mika Hämäläinen}
\email{mika.hamalainen@helsinki.fi}
\affiliation{%
  \institution{University of Helsinki}
  \city{Helsinki}
  \country{Finland}
}

\author{Khalid Alnajjar}
\email{khalid.alnajjar@helsinki.fi}
\affiliation{%
  \institution{University of Helsinki}
  \city{Helsinki}
  \country{Finland}
}

\author{Thierry Poibeau}
\email{thierry.poibeau@ens.psl.eu}
\affiliation{%
  \institution{LATTICE, École Normale Supérieure}
  \city{Paris}
  \country{France}
}

\renewcommand{\shortauthors}{Hämäläinen, Alnajjar and Poibeau}

\begin{abstract}
We present a method for extracting a multilingual sentiment annotated dialog data set from Fallout New Vegas. The game developers have preannotated every line of dialog in the game in one of the 8 different sentiments: \textit{anger, disgust, fear, happy, neutral, pained, sad } and \textit{surprised}. The game has been translated into English, Spanish, German, French and Italian. We conduct experiments on multilingual, multilabel sentiment analysis on the extracted data set using multilingual BERT, XLMRoBERTa and language specific BERT models. In our experiments, multilingual BERT outperformed XLMRoBERTa for most of the languages, also language specific models were slightly better than multilingual BERT for most of the languages. The best overall accuracy was 54\% and it was achieved by using multilingual BERT on Spanish data. The extracted data set presents a challenging task for sentiment analysis. We have released the data, including the testing and training splits, openly on Zenodo. The data set has been shuffled for copyright reasons.
\end{abstract}

\ccsdesc[500]{Computing methodologies~Natural language processing}
\ccsdesc[500]{Computing methodologies~Natural language generation}
\ccsdesc[300]{Computing methodologies~Neural networks}

\keywords{sentiment analysis, video games as corpus, multilinguality}


\maketitle

\section{Introduction}

Multilabel sentiment analysis is a challenging NLP task. Sentiment analysis is often conducted in a simplified positive-negative scale in the field of NLP with highly successful results \cite{yang-etal-2021-exploring,yang-etal-2021-multimodal,barnes-etal-2021-structured,alnajjar2021word}. What still remains difficult is when a model needs to predict a more nuanced sentiment label such as sadness, happiness or anger. In this paper, we explore the latter case where sentiment is not reduced into one axis but rather annotated based on 8 different emotions.

Role playing games (RPGs) contain large amounts of dialog that could be used in a variety of NLP tasks. However, thus far, the resources RPGs have have been mostly gone unnoticed in the NLP research community. The video game we base our research on in this paper, Fallout New Vegas\footnote{https://fallout.bethesda.net/en/games/fallout-new-vegas}, contains a very interesting annotated dataset. The Steam version of the game is available in English, Spanish, French, Italian and German and each line of dialogue in the game has been annotated for sentiment and its intensity.

Fallout New Vegas is an open-world RPG that is set in a post-apocalyptic world after a nuclear attack. The game mixes RPG and first person shooter genres. The game has several side quests and many non-playable characters (NPCs) with long prescripted dialog containing multiple dialog options. All in all, the game contains over 53,000 lines of dialog making it an ideal corpus for a variety of NLP tasks.

In the current work, we describe how we extracted the preannotated dialog from Fallout New Vegas and we present several BERT-based models for sentiment analysis. We have released the extracted dataset with train-test splits openly on Zenodo\footnote{https://zenodo.org/record/6990638}. The data is of a high value because it represents a professionally annotated data source in 5 languages and in a completely new domain of text than the existing sentiment analysis corpora. We also showcase that video games are an extremely useful data source for NLP and their use as a corpus should be more widely studied. This research opens up a possibility to better understand and model emotion in dialog with computational methods.

\section{Related work}

In this section, we will focus on some of the research related to using video games as a resource for NLP tasks and also the latest advances in multilabel sentiment analysis. 

\begin{table*}[!ht]
\centering
\footnotesize
\begin{tabular}{|l|l|}
\hline
file name & VDialogueA\_VDialogueArcade\_00163394\_2                                                                                                                                                 \\ \hline
sentiment & Anger                                                                                                                                                                                    \\ \hline
English   & \begin{tabular}[c]{@{}l@{}}I'm just saying that if it were to fall into Lake Mead and be irreparably damaged...\\ and if you threw an EMP grenade in after it...\end{tabular}            \\ \hline
Spanish   & \begin{tabular}[c]{@{}l@{}}Solo digo que si se cayera al lago Mead y sufriera daños irreparables... \\ y luego tú le tiraras una granada...\end{tabular}                                 \\ \hline
French    & \begin{tabular}[c]{@{}l@{}}Je dis juste que s'il devait "tomber" dans le Lac Mead et souffrir de pannes irréversibles... \\ et qu'en plus vous lui lanciez une grenade IEM…\end{tabular} \\ \hline
German    & \begin{tabular}[c]{@{}l@{}}Was ich sagen will, ist, dass er ja durchaus in den Lake Meade "fallen" könnte ... \\ und wenn Sie eine EMP-Granate hinterherwerfen würden ...\end{tabular}   \\ \hline
Italian   & \begin{tabular}[c]{@{}l@{}}Dico solo che se dovesse cadere in Lake Mead e rimanere irreparabilmente danneggiato... \\ e se gli gettassi dietro anche una granata IEM...\end{tabular}     \\ \hline
\end{tabular}
\caption{An aligned example of one sentence in all the 5 languages}
\label{tab:examplesent}
\end{table*}

Video games have been used as a corpus before, for example, a recent paper extracts a sentiment lexicon from Skyrim \cite{bergsma2020creating}. Their approach is different from us in the sense that their focus was to extract a lexicon, where as our approach extracts sentiment annotated sentences to be used as training data for a machine learning model. Sentiment lexica play a very different role in sentiment analysis than machine learning approaches \cite{emilysentiment}. Fallout 4 dialog has been used before to as data for a dialog adaptation model \cite{10.1145/3337722.3341865}. The paper presents a machine learning approach for adapting existing dialog in Fallout 4 to better match the condition the game character is in. Also, video game data has been used to detect persuasion  \cite{poyhonen2022multilingual}.

Multilabel sentiment analysis is a fragmented field of research in the sense that several different approaches deal with different number of sentiment labels and the labels themselves can be higher level emotions (like in our case) or lower level affects. The term multilabel itself can also be understood differently, for instance there is research that calls identifying multiple sentiments (postive-negative) within a sentence multilabel sentiment analysis \cite{tao2020toward}. However, for us, multilabel means that there are more than the typical positve-negative axis to be considered in sentiment analysis.

A recent paper demonstrates multilabel sentiment analysis on 100 languages \cite{yilmaz2021multi}. The authors use RoBERTa-XLM to extract feature vectors. These are then used in training a bi-directional LSTM based classifier model. Another line of work \cite{liu2015multi} compares several different multilabel classification methods on the task of sentiment analysis showing that RAkEL \cite{tsoumakas2010random} gave the best performance on raw token input. They show promising results using both of the models but conclude that multilabel sentiment analysis is far from an easy task using machine learning methods.

\section{Extracting the data}

Fallout New Vegas stores data in a proprietary binary format that cannot be parsed easily without specialized tools. Fortunately, the developers have released an official modding tool called Garden of Eden Creation Kit (GECK)\footnote{https://geck.bethsoft.com/index.php?title=Main\_Page} that allows us to extract all in-game dialog. We use the version of the game that is distributed through Steam\footnote{https://store.steampowered.com/app/22380/Fallout\_New \_Vegas/}. GECK outputs a TSV file out of which we save the text, file name and sentiment. We use the file name information to align sentences in different languages with each other. And example of the resulting data in the different languages can be seen in Table \ref{tab:examplesent}.

\begin{table}[ht]
\centering
\footnotesize
\begin{tabular}{|l|l|l|l|l|l|}
\hline
          & Eng & Deu & Ita & Spa & Fra \\ \hline
Anger     & 3335    & 2407   & 2189    & 611     & 143    \\ \hline
Disgust   & 932     & 679    & 658     & 206     & 34     \\ \hline
Fear      & 1620    & 969    & 726     & 140     & 34     \\ \hline
Happy     & 4029    & 2351   & 1916    & 375     & 103    \\ \hline
Neutral   & 39802   & 29339  & 25664   & 5819    & 1626   \\ \hline
Pained    & 994     & 846    & 780     & 89      & 12     \\ \hline
Sad       & 1055    & 714    & 649     & 248     & 64     \\ \hline
Surprised & 1649    & 1082   & 930     & 167     & 59     \\ \hline
\end{tabular}
\caption{Number of dialog lines per language and sentiment label}
\label{tab:datasize}
\end{table}

The problem we ran into was that GECK is rather buggy, we tried to patch it using an unofficial tool called GECK Extender\footnote{https://geckwiki.com/index.php/GECK\_Extender}, but it did not patch the main bug we encountered. GECK manages to export the entire in-game dialog correctly for English, but it crashes after some time for other languages. It seems to crash consistently at the same step for a given language. For German and Italian, it extracts majority of the dialog before crashing, where as for Spanish and French, it crashes rather early in the extraction process. We also tried out some unofficial tools such as fo76utils\footnote{https://github.com/fo76utils/fo76utils} but they did not work at all with Fallout New Vegas. As a result, we have slightly different amount of data for each language as seen in Table \ref{tab:datasize}.

Because the data is so small for Spanish and French, we will not use them for training. Instead, we use the sentences from the Spanish and French data for testing. We also exclude the translations of these sentences in other languages from the training and use them for testing. This way all languages share the testing data and no translation of the test sentences appears in training for any of the languages. The rest of the data for English, German and Italian is used for training. Because the Neutral label is so overwhelmingly present in the data, we limit it to 3000 samples in the training data for each language.

The data itself is relatively clean because it is dialog displayed to the end user in the video game. Some of the dialog lines include additional annotation inside of brackets: \textit{\{nervous, hiding a secret\}Corporal White? I don't know where he - \{obviously changing his lie mid-sentence\}uh, I mean, never heard of him. Uh, I gotta go...} We removed these extra annotations with regular expressions. Apart from this, no additional preprocessing was done.

\begin{table*}[ht]
\centering
\setlength{\tabcolsep}{0.15em}
\resizebox{15cm}{!}{%
\begin{tabular}{|l|llllll|llllll|llllll|llllll|llllll|}
\hline
 & \multicolumn{6}{c|}{English} & \multicolumn{6}{c|}{German} & \multicolumn{6}{c|}{Italian} & \multicolumn{6}{c|}{Spanish} & \multicolumn{6}{c|}{French} \\ \hline
model & \multicolumn{3}{c|}{BERT} & \multicolumn{3}{c|}{ROBERTA} & \multicolumn{3}{c|}{BERT} & \multicolumn{3}{c|}{ROBERTA} & \multicolumn{3}{c|}{BERT} & \multicolumn{3}{c|}{ROBERTA} & \multicolumn{3}{c|}{BERT} & \multicolumn{3}{l|}{ROBERTA} & \multicolumn{3}{c|}{BERT} & \multicolumn{3}{c|}{ROBERTA} \\ \hline
 & \multicolumn{1}{l|}{pre} & \multicolumn{1}{l|}{\cellcolor[HTML]{EFEFEF}rec} & \multicolumn{1}{l|}{f1} & \multicolumn{1}{l|}{\cellcolor[HTML]{EFEFEF}pre} & \multicolumn{1}{l|}{rec} & \cellcolor[HTML]{EFEFEF}f1 & \multicolumn{1}{l|}{pre} & \multicolumn{1}{l|}{\cellcolor[HTML]{EFEFEF}rec} & \multicolumn{1}{l|}{f1} & \multicolumn{1}{l|}{\cellcolor[HTML]{EFEFEF}pre} & \multicolumn{1}{l|}{rec} & \cellcolor[HTML]{EFEFEF}f1 & \multicolumn{1}{l|}{pre} & \multicolumn{1}{l|}{\cellcolor[HTML]{EFEFEF}rec} & \multicolumn{1}{l|}{f1} & \multicolumn{1}{l|}{\cellcolor[HTML]{EFEFEF}pre} & \multicolumn{1}{l|}{rec} & \cellcolor[HTML]{EFEFEF}f1 & \multicolumn{1}{l|}{pre} & \multicolumn{1}{l|}{\cellcolor[HTML]{EFEFEF}rec} & \multicolumn{1}{l|}{f1} & \multicolumn{1}{l|}{\cellcolor[HTML]{EFEFEF}pre} & \multicolumn{1}{l|}{rec} & \cellcolor[HTML]{EFEFEF}f1 & \multicolumn{1}{l|}{pre} & \multicolumn{1}{l|}{\cellcolor[HTML]{EFEFEF}rec} & \multicolumn{1}{l|}{f1} & \multicolumn{1}{l|}{\cellcolor[HTML]{EFEFEF}pre} & \multicolumn{1}{l|}{rec} & \cellcolor[HTML]{EFEFEF}f1 \\ \hline
Anger & \multicolumn{1}{l|}{0.16} & \multicolumn{1}{l|}{\cellcolor[HTML]{EFEFEF}0.29} & \multicolumn{1}{l|}{0.21} & \multicolumn{1}{l|}{\cellcolor[HTML]{EFEFEF}0.18} & \multicolumn{1}{l|}{0.30} & \cellcolor[HTML]{EFEFEF}0.22 & \multicolumn{1}{l|}{0.14} & \multicolumn{1}{l|}{\cellcolor[HTML]{EFEFEF}0.23} & \multicolumn{1}{l|}{0.17} & \multicolumn{1}{l|}{\cellcolor[HTML]{EFEFEF}0.15} & \multicolumn{1}{l|}{0.22} & \cellcolor[HTML]{EFEFEF}0.17 & \multicolumn{1}{l|}{0.14} & \multicolumn{1}{l|}{\cellcolor[HTML]{EFEFEF}0.23} & \multicolumn{1}{l|}{0.18} & \multicolumn{1}{l|}{\cellcolor[HTML]{EFEFEF}0.17} & \multicolumn{1}{l|}{0.25} & \cellcolor[HTML]{EFEFEF}0.20 & \multicolumn{1}{l|}{0.12} & \multicolumn{1}{l|}{\cellcolor[HTML]{EFEFEF}0.15} & \multicolumn{1}{l|}{0.13} & \multicolumn{1}{l|}{\cellcolor[HTML]{EFEFEF}0.16} & \multicolumn{1}{l|}{0.19} & \cellcolor[HTML]{EFEFEF}0.17 & \multicolumn{1}{l|}{0.13} & \multicolumn{1}{l|}{\cellcolor[HTML]{EFEFEF}0.29} & \multicolumn{1}{l|}{0.18} & \multicolumn{1}{l|}{\cellcolor[HTML]{EFEFEF}0.16} & \multicolumn{1}{l|}{0.22} & \cellcolor[HTML]{EFEFEF}0.18 \\ \hline
Disgust & \multicolumn{1}{l|}{0.07} & \multicolumn{1}{l|}{\cellcolor[HTML]{EFEFEF}0.11} & \multicolumn{1}{l|}{0.09} & \multicolumn{1}{l|}{\cellcolor[HTML]{EFEFEF}0.11} & \multicolumn{1}{l|}{0.11} & \cellcolor[HTML]{EFEFEF}0.11 & \multicolumn{1}{l|}{0.09} & \multicolumn{1}{l|}{\cellcolor[HTML]{EFEFEF}0.12} & \multicolumn{1}{l|}{0.10} & \multicolumn{1}{l|}{\cellcolor[HTML]{EFEFEF}0.11} & \multicolumn{1}{l|}{0.11} & \cellcolor[HTML]{EFEFEF}0.11 & \multicolumn{1}{l|}{0.10} & \multicolumn{1}{l|}{\cellcolor[HTML]{EFEFEF}0.11} & \multicolumn{1}{l|}{0.10} & \multicolumn{1}{l|}{\cellcolor[HTML]{EFEFEF}0.11} & \multicolumn{1}{l|}{0.10} & \cellcolor[HTML]{EFEFEF}0.11 & \multicolumn{1}{l|}{0.04} & \multicolumn{1}{l|}{\cellcolor[HTML]{EFEFEF}0.02} & \multicolumn{1}{l|}{0.03} & \multicolumn{1}{l|}{\cellcolor[HTML]{EFEFEF}0.07} & \multicolumn{1}{l|}{0.05} & \cellcolor[HTML]{EFEFEF}0.06 & \multicolumn{1}{l|}{0.09} & \multicolumn{1}{l|}{\cellcolor[HTML]{EFEFEF}0.15} & \multicolumn{1}{l|}{0.11} & \multicolumn{1}{l|}{\cellcolor[HTML]{EFEFEF}0.00} & \multicolumn{1}{l|}{0.00} & \cellcolor[HTML]{EFEFEF}0.00 \\ \hline
Fear & \multicolumn{1}{l|}{0.17} & \multicolumn{1}{l|}{\cellcolor[HTML]{EFEFEF}0.29} & \multicolumn{1}{l|}{0.22} & \multicolumn{1}{l|}{\cellcolor[HTML]{EFEFEF}0.13} & \multicolumn{1}{l|}{0.28} & \cellcolor[HTML]{EFEFEF}0.17 & \multicolumn{1}{l|}{0.05} & \multicolumn{1}{l|}{\cellcolor[HTML]{EFEFEF}0.06} & \multicolumn{1}{l|}{0.06} & \multicolumn{1}{l|}{\cellcolor[HTML]{EFEFEF}0.08} & \multicolumn{1}{l|}{0.13} & \cellcolor[HTML]{EFEFEF}0.10 & \multicolumn{1}{l|}{0.03} & \multicolumn{1}{l|}{\cellcolor[HTML]{EFEFEF}0.03} & \multicolumn{1}{l|}{0.03} & \multicolumn{1}{l|}{\cellcolor[HTML]{EFEFEF}0.10} & \multicolumn{1}{l|}{0.18} & \cellcolor[HTML]{EFEFEF}0.13 & \multicolumn{1}{l|}{0.07} & \multicolumn{1}{l|}{\cellcolor[HTML]{EFEFEF}0.09} & \multicolumn{1}{l|}{0.08} & \multicolumn{1}{l|}{\cellcolor[HTML]{EFEFEF}0.06} & \multicolumn{1}{l|}{0.10} & \cellcolor[HTML]{EFEFEF}0.08 & \multicolumn{1}{l|}{0.14} & \multicolumn{1}{l|}{\cellcolor[HTML]{EFEFEF}0.12} & \multicolumn{1}{l|}{0.13} & \multicolumn{1}{l|}{\cellcolor[HTML]{EFEFEF}0.12} & \multicolumn{1}{l|}{0.21} & \cellcolor[HTML]{EFEFEF}0.15 \\ \hline
Happy & \multicolumn{1}{l|}{0.11} & \multicolumn{1}{l|}{\cellcolor[HTML]{EFEFEF}0.28} & \multicolumn{1}{l|}{0.16} & \multicolumn{1}{l|}{\cellcolor[HTML]{EFEFEF}0.12} & \multicolumn{1}{l|}{0.38} & \cellcolor[HTML]{EFEFEF}0.19 & \multicolumn{1}{l|}{0.08} & \multicolumn{1}{l|}{\cellcolor[HTML]{EFEFEF}0.18} & \multicolumn{1}{l|}{0.11} & \multicolumn{1}{l|}{\cellcolor[HTML]{EFEFEF}0.11} & \multicolumn{1}{l|}{0.33} & \cellcolor[HTML]{EFEFEF}0.16 & \multicolumn{1}{l|}{0.10} & \multicolumn{1}{l|}{\cellcolor[HTML]{EFEFEF}0.28} & \multicolumn{1}{l|}{0.14} & \multicolumn{1}{l|}{\cellcolor[HTML]{EFEFEF}0.11} & \multicolumn{1}{l|}{0.32} & \cellcolor[HTML]{EFEFEF}0.16 & \multicolumn{1}{l|}{0.09} & \multicolumn{1}{l|}{\cellcolor[HTML]{EFEFEF}0.23} & \multicolumn{1}{l|}{0.13} & \multicolumn{1}{l|}{\cellcolor[HTML]{EFEFEF}0.12} & \multicolumn{1}{l|}{0.39} & \cellcolor[HTML]{EFEFEF}0.18 & \multicolumn{1}{l|}{0.07} & \multicolumn{1}{l|}{\cellcolor[HTML]{EFEFEF}0.20} & \multicolumn{1}{l|}{0.10} & \multicolumn{1}{l|}{\cellcolor[HTML]{EFEFEF}0.10} & \multicolumn{1}{l|}{0.31} & \cellcolor[HTML]{EFEFEF}0.15 \\ \hline
Neutral & \multicolumn{1}{l|}{0.82} & \multicolumn{1}{l|}{\cellcolor[HTML]{EFEFEF}0.59} & \multicolumn{1}{l|}{0.69} & \multicolumn{1}{l|}{\cellcolor[HTML]{EFEFEF}0.82} & \multicolumn{1}{l|}{0.57} & \cellcolor[HTML]{EFEFEF}0.67 & \multicolumn{1}{l|}{0.80} & \multicolumn{1}{l|}{\cellcolor[HTML]{EFEFEF}0.63} & \multicolumn{1}{l|}{0.70} & \multicolumn{1}{l|}{\cellcolor[HTML]{EFEFEF}0.81} & \multicolumn{1}{l|}{0.61} & \cellcolor[HTML]{EFEFEF}0.69 & \multicolumn{1}{l|}{0.81} & \multicolumn{1}{l|}{\cellcolor[HTML]{EFEFEF}0.62} & \multicolumn{1}{l|}{0.70} & \multicolumn{1}{l|}{\cellcolor[HTML]{EFEFEF}0.81} & \multicolumn{1}{l|}{0.62} & \cellcolor[HTML]{EFEFEF}0.70 & \multicolumn{1}{l|}{0.79} & \multicolumn{1}{l|}{\cellcolor[HTML]{EFEFEF}0.67} & \multicolumn{1}{l|}{0.72} & \multicolumn{1}{l|}{\cellcolor[HTML]{EFEFEF}0.80} & \multicolumn{1}{l|}{0.63} & \cellcolor[HTML]{EFEFEF}0.70 & \multicolumn{1}{l|}{0.83} & \multicolumn{1}{l|}{\cellcolor[HTML]{EFEFEF}0.58} & \multicolumn{1}{l|}{0.68} & \multicolumn{1}{l|}{\cellcolor[HTML]{EFEFEF}0.82} & \multicolumn{1}{l|}{0.64} & \cellcolor[HTML]{EFEFEF}0.72 \\ \hline
Pained & \multicolumn{1}{l|}{0.08} & \multicolumn{1}{l|}{\cellcolor[HTML]{EFEFEF}0.08} & \multicolumn{1}{l|}{0.08} & \multicolumn{1}{l|}{\cellcolor[HTML]{EFEFEF}0.08} & \multicolumn{1}{l|}{0.07} & \cellcolor[HTML]{EFEFEF}0.07 & \multicolumn{1}{l|}{0.04} & \multicolumn{1}{l|}{\cellcolor[HTML]{EFEFEF}0.03} & \multicolumn{1}{l|}{0.04} & \multicolumn{1}{l|}{\cellcolor[HTML]{EFEFEF}0.03} & \multicolumn{1}{l|}{0.03} & \cellcolor[HTML]{EFEFEF}0.03 & \multicolumn{1}{l|}{0.03} & \multicolumn{1}{l|}{\cellcolor[HTML]{EFEFEF}0.03} & \multicolumn{1}{l|}{0.03} & \multicolumn{1}{l|}{\cellcolor[HTML]{EFEFEF}0.04} & \multicolumn{1}{l|}{0.03} & \cellcolor[HTML]{EFEFEF}0.04 & \multicolumn{1}{l|}{0.04} & \multicolumn{1}{l|}{\cellcolor[HTML]{EFEFEF}0.03} & \multicolumn{1}{l|}{0.04} & \multicolumn{1}{l|}{\cellcolor[HTML]{EFEFEF}0.03} & \multicolumn{1}{l|}{0.02} & \cellcolor[HTML]{EFEFEF}0.02 & \multicolumn{1}{l|}{0.00} & \multicolumn{1}{l|}{\cellcolor[HTML]{EFEFEF}0.00} & \multicolumn{1}{l|}{0.00} & \multicolumn{1}{l|}{\cellcolor[HTML]{EFEFEF}0.00} & \multicolumn{1}{l|}{0.00} & \cellcolor[HTML]{EFEFEF}0.00 \\ \hline
Sad & \multicolumn{1}{l|}{0.10} & \multicolumn{1}{l|}{\cellcolor[HTML]{EFEFEF}0.14} & \multicolumn{1}{l|}{0.12} & \multicolumn{1}{l|}{\cellcolor[HTML]{EFEFEF}0.10} & \multicolumn{1}{l|}{0.14} & \cellcolor[HTML]{EFEFEF}0.12 & \multicolumn{1}{l|}{0.09} & \multicolumn{1}{l|}{\cellcolor[HTML]{EFEFEF}0.11} & \multicolumn{1}{l|}{0.10} & \multicolumn{1}{l|}{\cellcolor[HTML]{EFEFEF}0.09} & \multicolumn{1}{l|}{0.10} & \cellcolor[HTML]{EFEFEF}0.09 & \multicolumn{1}{l|}{0.09} & \multicolumn{1}{l|}{\cellcolor[HTML]{EFEFEF}0.11} & \multicolumn{1}{l|}{0.10} & \multicolumn{1}{l|}{\cellcolor[HTML]{EFEFEF}0.10} & \multicolumn{1}{l|}{0.13} & \cellcolor[HTML]{EFEFEF}0.12 & \multicolumn{1}{l|}{0.10} & \multicolumn{1}{l|}{\cellcolor[HTML]{EFEFEF}0.10} & \multicolumn{1}{l|}{0.10} & \multicolumn{1}{l|}{\cellcolor[HTML]{EFEFEF}0.10} & \multicolumn{1}{l|}{0.10} & \cellcolor[HTML]{EFEFEF}0.10 & \multicolumn{1}{l|}{0.04} & \multicolumn{1}{l|}{\cellcolor[HTML]{EFEFEF}0.08} & \multicolumn{1}{l|}{0.06} & \multicolumn{1}{l|}{\cellcolor[HTML]{EFEFEF}0.12} & \multicolumn{1}{l|}{0.14} & \cellcolor[HTML]{EFEFEF}0.13 \\ \hline
Surprised & \multicolumn{1}{l|}{0.08} & \multicolumn{1}{l|}{\cellcolor[HTML]{EFEFEF}0.18} & \multicolumn{1}{l|}{0.11} & \multicolumn{1}{l|}{\cellcolor[HTML]{EFEFEF}0.07} & \multicolumn{1}{l|}{0.19} & \cellcolor[HTML]{EFEFEF}0.10 & \multicolumn{1}{l|}{0.09} & \multicolumn{1}{l|}{\cellcolor[HTML]{EFEFEF}0.17} & \multicolumn{1}{l|}{0.12} & \multicolumn{1}{l|}{\cellcolor[HTML]{EFEFEF}0.07} & \multicolumn{1}{l|}{0.17} & \cellcolor[HTML]{EFEFEF}0.10 & \multicolumn{1}{l|}{0.06} & \multicolumn{1}{l|}{\cellcolor[HTML]{EFEFEF}0.13} & \multicolumn{1}{l|}{0.09} & \multicolumn{1}{l|}{\cellcolor[HTML]{EFEFEF}0.09} & \multicolumn{1}{l|}{0.20} & \cellcolor[HTML]{EFEFEF}0.12 & \multicolumn{1}{l|}{0.06} & \multicolumn{1}{l|}{\cellcolor[HTML]{EFEFEF}0.14} & \multicolumn{1}{l|}{0.09} & \multicolumn{1}{l|}{\cellcolor[HTML]{EFEFEF}0.08} & \multicolumn{1}{l|}{0.17} & \cellcolor[HTML]{EFEFEF}0.11 & \multicolumn{1}{l|}{0.04} & \multicolumn{1}{l|}{\cellcolor[HTML]{EFEFEF}0.07} & \multicolumn{1}{l|}{0.05} & \multicolumn{1}{l|}{\cellcolor[HTML]{EFEFEF}0.14} & \multicolumn{1}{l|}{0.25} & \cellcolor[HTML]{EFEFEF}0.18 \\ \hline
\begin{tabular}[c]{@{}l@{}}Overall\\ accuracy\end{tabular} & \multicolumn{3}{c|}{\textbf{0.51}} & \multicolumn{3}{c|}{0.50} & \multicolumn{3}{c|}{\textbf{0.52}} & \multicolumn{3}{c|}{0.51} & \multicolumn{3}{c|}{\textbf{0.52}} & \multicolumn{3}{c|}{\textbf{0.52}} & \multicolumn{3}{c|}{\textbf{0.54}} & \multicolumn{3}{c|}{0.52} & \multicolumn{3}{c|}{0.49} & \multicolumn{3}{c|}{\textbf{0.55}} \\ \hline
\end{tabular}%
}
\caption{Results of the multilingual models (precision, recall, F1-score)}
\label{tab:multilingresults}
\end{table*}

\section{Sentiment analysis}

We model the sentiment analysis task as a sequence classification task, where the model has to predict the sentiment label given a Fallout New Vegas dialog sentence. We experiment with several BERT-based \cite{devlin-etal-2019-bert} models and a RoBERTa-based \cite{Liu2019RoBERTaAR} model. We experiment both in a multilingual and a monolingual scenario. We split  15\% of the training data for validation for each trainable language.

\subsection{Multilingual setting}

In our multilingual models, the model is trained with the training data for English, Italian and German. Then the models are evaluated using the evaluation splits for English, Italian and German, and the entire data set for Spanish and French which do not have enough data for training.

We use the transformers Pyton library \cite{wolf-etal-2020-transformers} to fine-tune the multilingual BERT \cite{devlin-etal-2019-bert}\footnote{bert-base-multilingual-cased} and the multilingual XLMRoBERTa \cite{conneau-etal-2020-unsupervised}\footnote{xlm-roberta-base} for sequence classification task. Both of the models are trained with the same data for 3 epochs with 500 warm-up steps and a weight decay of 0.01. The labels are predicted using the softmax function.

Because both of the models are multilingual they should learn to predict the sentiment even for Spanish and French which were completely held out from the training data. Because we intend to test the models on these two languages, we do not introduce any language labels to distinguish between languages during the training.

\subsection{Monolingual setting}

In addition to training multilingual models with the data for all available languages, we are interested in seeing whether monolingual models perform better or worse in this task than the multilingual model. Monolingual models are usually trained with more data for the particular language in question than what is present in the multilingual models. We train separate models for English, Italian and German, training and testing them only with language specific data.

For English, we use original BERT model \cite{devlin-etal-2019-bert}\footnote{bert-base-uncased}. For Italian, we use the Italian BERT\footnote{dbmdz/bert-base-italian-uncased} provided by the MDZ Digital Library team (dbmdz) at the Bavarian State Library. The model is trained on OPUS and OSCAR corpora. The German BERT model\footnote{bert-base-german-dbmdz-uncased} is also provided by the same team and it has been trained on a variety of corpora such as a Wikipedia dump, EU Bookshop corpus, Open Subtitles, CommonCrawl, ParaCrawl and News Crawl. We use the same training parameters as for the multilingual models.

\section{Results and evaluation}

The results of the multilingual models can be seen in Table \ref{tab:multilingresults}. The results are calculated using the \textit{classification\_report} method of Scikit-learn \cite{scikit-learn}. There is not a big difference between the multilingual BERT and XLMRoBERTa models. XLMRoBERTa seems to be slightly worse than the multilingual BERT, however, it has a better overall accuracy for French. All in all, the sentiments that are the most difficult ones to predict seem to be \textit{disgust} and \textit{pained}. Perhaps because they might depend more on the audio cues than textual ones. After the \textit{neutral} label, \textit{anger} and \textit{happy} seem to be the easiest ones to predict for the model, although the overall performance is not very good.

We can see that for both of the multilingual models could learn to transfer the sentiment analysis to Spanish and French, which were not used in the training at all. The overall accuracy is comparable to that of the languages that had training data, however, the results for some labels are rather poor. For instance, for French, neither of the models predicted any \textit{pained} sentiment sentences right.


The results for the monolingual models can be seen in Table \ref{tab:monolingresults}. The results are slightly better for all languages than Italian when comparing to the multilingual models. However, there is no major increase in the overall accuracy either despite the models having been pretrained with more language specific data. We believe that this is due to the fact that Fallout belongs to a very different domain of text than what is represented by the training data of the BERT models.

\begin{table}[]
\centering
\footnotesize
\setlength{\tabcolsep}{0.15em}
\begin{tabular}{|l|lll|lll|lll|}
\hline
                                                           & \multicolumn{3}{c|}{English}                                                         & \multicolumn{3}{c|}{German}                                                          & \multicolumn{3}{c|}{Italian}                                                         \\ \hline
                                                           & \multicolumn{1}{l|}{pre}  & \multicolumn{1}{l|}{rec}                          & f1   & \multicolumn{1}{l|}{pre}  & \multicolumn{1}{l|}{rec}                          & f1   & \multicolumn{1}{l|}{pre}  & \multicolumn{1}{l|}{rec}                          & f1   \\ \hline
Anger                                                      & \multicolumn{1}{l|}{0.22} & \multicolumn{1}{l|}{\cellcolor[HTML]{EFEFEF}0.31} & 0.26 & \multicolumn{1}{l|}{0.18} & \multicolumn{1}{l|}{\cellcolor[HTML]{EFEFEF}0.28} & 0.22 & \multicolumn{1}{l|}{0.17} & \multicolumn{1}{l|}{\cellcolor[HTML]{EFEFEF}0.30} & 0.22 \\ \hline
Disgust                                                    & \multicolumn{1}{l|}{0.12} & \multicolumn{1}{l|}{\cellcolor[HTML]{EFEFEF}0.18} & 0.15 & \multicolumn{1}{l|}{0.09} & \multicolumn{1}{l|}{\cellcolor[HTML]{EFEFEF}0.13} & 0.11 & \multicolumn{1}{l|}{0.07} & \multicolumn{1}{l|}{\cellcolor[HTML]{EFEFEF}0.11} & 0.09 \\ \hline
Fear                                                       & \multicolumn{1}{l|}{0.17} & \multicolumn{1}{l|}{\cellcolor[HTML]{EFEFEF}0.32} & 0.23 & \multicolumn{1}{l|}{0.04} & \multicolumn{1}{l|}{\cellcolor[HTML]{EFEFEF}0.03} & 0.03 & \multicolumn{1}{l|}{0.04} & \multicolumn{1}{l|}{\cellcolor[HTML]{EFEFEF}0.06} & 0.05 \\ \hline
Happy                                                      & \multicolumn{1}{l|}{0.14} & \multicolumn{1}{l|}{\cellcolor[HTML]{EFEFEF}0.41} & 0.21 & \multicolumn{1}{l|}{0.12} & \multicolumn{1}{l|}{\cellcolor[HTML]{EFEFEF}0.31} & 0.17 & \multicolumn{1}{l|}{0.12} & \multicolumn{1}{l|}{\cellcolor[HTML]{EFEFEF}0.29} & 0.17 \\ \hline
Neutral                                                    & \multicolumn{1}{l|}{0.82} & \multicolumn{1}{l|}{\cellcolor[HTML]{EFEFEF}0.60} & 0.69 & \multicolumn{1}{l|}{0.81} & \multicolumn{1}{l|}{\cellcolor[HTML]{EFEFEF}0.64} & 0.72 & \multicolumn{1}{l|}{0.81} & \multicolumn{1}{l|}{\cellcolor[HTML]{EFEFEF}0.58} & 0.68 \\ \hline
Pained                                                     & \multicolumn{1}{l|}{0.05} & \multicolumn{1}{l|}{\cellcolor[HTML]{EFEFEF}0.09} & 0.06 & \multicolumn{1}{l|}{0.03} & \multicolumn{1}{l|}{\cellcolor[HTML]{EFEFEF}0.06} & 0.04 & \multicolumn{1}{l|}{0.03} & \multicolumn{1}{l|}{\cellcolor[HTML]{EFEFEF}0.03} & 0.03 \\ \hline
Sad                                                        & \multicolumn{1}{l|}{0.11} & \multicolumn{1}{l|}{\cellcolor[HTML]{EFEFEF}0.13} & 0.12 & \multicolumn{1}{l|}{0.11} & \multicolumn{1}{l|}{\cellcolor[HTML]{EFEFEF}0.13} & 0.12 & \multicolumn{1}{l|}{0.11} & \multicolumn{1}{l|}{\cellcolor[HTML]{EFEFEF}0.20} & 0.14 \\ \hline
Surprised                                                  & \multicolumn{1}{l|}{0.10} & \multicolumn{1}{l|}{\cellcolor[HTML]{EFEFEF}0.25} & 0.14 & \multicolumn{1}{l|}{0.08} & \multicolumn{1}{l|}{\cellcolor[HTML]{EFEFEF}0.17} & 0.11 & \multicolumn{1}{l|}{0.08} & \multicolumn{1}{l|}{\cellcolor[HTML]{EFEFEF}0.18} & 0.11 \\ \hline
\begin{tabular}[c]{@{}l@{}}Overall\\ accuracy\end{tabular} & \multicolumn{3}{c|}{0.52}                                                            & \multicolumn{3}{c|}{0.54}                                                            & \multicolumn{3}{c|}{0.50}                                                            \\ \hline
\end{tabular}
\caption{Results of the language specific models}
\label{tab:monolingresults}
\end{table}

Next, we will take a closer look at the results of the multilingual BERT model as it seemed to give better results than the XLMRoBERTa based one. If we look at the aligned sentences that were predicted wrong for all of the languages, we can see that there are 269 such cases. On a label level, the model predicted the same sentence wrong for all languages 29 times for \textit{surprise}, 30 for \textit{happy}, 11 for \textit{fear}, 46 for \textit{anger}, 103 for \textit{neutral}, 31 for \textit{sad}, 8 for \textit{pained} and 11 for \textit{disgust} label. It is interesting that the model did not commit the same errors for all languages even though the sentences are translations of each other, and if the model was to truly capture the multilinguality of semantics, it ought to make same mistake for sentences that are each other's translations.

For example the following \textit{neutral} sentence was predicted either as \textit{sad} or \textit{disgust} depending on the language: \textit{Why the need for a bunch of old warhorses like us? }(\textit{disgust}), \textit{Woher rührt der Bedarf nach einem Haufen alter Haudegen wie uns?} (\textit{disgust}), \textit{Come mai il bisogno di un gruppo di vecchi veterani come noi?} (\textit{sad}), \textit{¿Para qué necesitas a un montón de veteranos decrépitos como nosotros?} (\textit{sad}) and \textit{Pourquoi faire appel à des vieux bourrins comme nous ?} (\textit{disgust}). The difference might be explained by the translation strategy used. The Spanish and Italian sentences use the word veteran where as the other languages use horse related vocabulary \textit{warhorse}, \textit{Haudegen} and \textit{bourrin}.

If we look at sentences that were predicted correctly for all of the languages, we can see 370 such cases. 363 of them are \textit{neutral}, 4 \textit{anger}, 2 \textit{surprise} and 1 \textit{happy}. This means that while the model gets the prediction right for at least one language, it seldom gets it fully right for all of the languages. For example, the following sentence was classified correctly as surprise by all of the models: \textit{Huh? Look, man, me and Diane, we don't dig on that politics stuff, savvy? We just make the product and make it get to a good home},  \textit{Häh? Ich und Diane, wir stehen nicht so auf Politikkram, ja? Wir stellen nur Ware her und sorgen dafür, dass sie ein ordentliches Zuhause bekommt}, \textit{Eh? Guarda, amico, io e Diane non ci occupiamo di politica, chiaro? Realizziamo solo il prodotto e facciamo sì che arrivi a destinazione}, \textit{¿Eh? Escucha, colega, Diane y yo no nos metemos en cosas de política, ¿entiendes? Solo fabricamos el producto y lo hacemos llegar a buen puerto} and \textit{Hein ? Écoutez, mec, Diane et moi, on n'est pas branchés politique, d'accord ? On fabrique les produits et on essaie d'en vivre le mieux possible}. In all of the cases, the translators had retained the initial \textit{huh?} which probably gave the model a good cue for predicting the label correctly as \textit{surprise}.

\section{Conclusions}

We have presented a new multilingual data set for sentiment analysis consisting of 8 sentiment labels. The data has been extracted from Fallout New Vegas which already contained sentiment labels. In addition to the labels, the game data contained sentiment intensity scores, but we did not utilize them in this research. We have made the data publicly available on Zenodo\footnote{https://zenodo.org/record/6990638}. The data has been shuffled for copyright reasons.

Multimodal classification of text remains a challenging NLP problem and our experiments on sentiment analysis are no exception. The overall accuracies of the models are rather low, but they are in line with the usual accuracies obtained in similar multilabel NLP classification tasks. Both multilingual BERT and XLMRoBERTa were able to learn to analyze sentiment in Spanish and French while they were excluded from the training.

In the future, we are interested in conducting sentiment analysis multimodally, because we believe that it would be helpful for many of the sentiments expressed in the corpus, as intonation and how the sentences are pronounced may contain better cues about the sentiment than pure text. It should be possible to get at least audio data because the game has voice acting. However, before that, we need to tackle the practical problem of actually extracting the audio files from the game. Currenlty, GECK outputs \textit{File not found} for the audio files of each sentence.

\begin{acks}
This work was partially financed by the Society of Swedish Literature in Finland with funding from Enhancing Conversational AI with Computational Creativity, and by the Ella and Georg Ehrnrooth Foundation for Modelling Conversational Artificial Intelligence with Intent and Creativity. This research has received mobility funding from Nokia Foundation under grant number 20220193.
\end{acks}

\bibliographystyle{ACM-Reference-Format}
\bibliography{sample-base}

\end{document}